\documentclass[sigconf,nonacm]{acmart}
\renewcommand\footnotetextcopyrightpermission[1]{}
\usepackage{booktabs}
\usepackage{pgfplotstable} 
\pgfplotsset{compat=1.18}
\usepackage{newunicodechar}
\usepackage[utf8]{inputenc}
\usepackage{enumitem}
\usepackage{booktabs}    
\usepackage{multirow}    
\usepackage{graphicx}    
\usepackage{amsmath}     
\AtBeginDocument{%
  }

\setcopyright{none}
\copyrightyear{2026}
\acmYear{2026}
\acmDOI{}

\begin{document}

\title{AIGS: Adaptive Incremental Gating System for Online Representation Learning in Non-Stationary Data Streams}



\author{SiRui He}
\affiliation{%
  \institution{Xiamen University Malaysia}
  \city{Sepang}
  \country{Malaysia}
}
\email{swe.hesirui@gmail.com}

\author{Kai Liang Lew}
\affiliation{%
  \institution{Faculty of Engineering \& Technology, Multimedia University}
  \streetaddress{Jalan Ayer Keroh Lama}
  \city{Melaka}
  \state{Melaka}
  \postcode{75450}
  \country{Malaysia}
}
\email{1132703002@student.mmu.edu.my}

\author{Chui Zi Ong}
\affiliation{%
  \institution{School of Economics and Management, Xiamen University Malaysia}
  \city{Sepang}
  \country{Malaysia}
}
\email{chuizi.ong@xmu.edu.my}

\author{Chean Khim Toa}
\authornote{Corresponding author.} 
\affiliation{%
  \institution{School of Computing and Data Science, Xiamen University Malaysia}
  \city{Sepang}
  \country{Malaysia}
}
\email{cheankhim.toa@xmu.edu.my}


\begin{abstract}
Real-time data streams in Web of Things (WoT) and edge computing environments often evolve through latent regime changes. For online representation learning under strict computational constraints, the central problem is resolving the stability-plasticity dilemma: keeping useful historical knowledge while rapidly reacting to concept drift. Existing methods employ fixed update schedules or rolling windows. However, they suffer from parameter ossification during sudden shifts and waste computational resources when the stream remains stable. This paper proposes the Adaptive Incremental Gating System (AIGS), a lightweight closed-loop state-aware adaptation framework. AIGS introduces the Shock Ratio, an endogenous residual feedback mechanism that normalizes current reconstruction error against recent variation. This signal drives a Continuous Plasticity Controller that smoothly interpolates between learning plasticity and memory retention. By treating representation learning as a closed-loop control mechanism, AIGS avoids catastrophic forgetting and maintains a strictly linear $\mathcal{O}\left(k\cdot d\right)$ per-step complexity suitable for latency-sensitive edge devices. Experiments on real-world smart city dynamic streams—spanning traffic networks, meteorological systems, and industrial infrastructure—demonstrate distinct domain-dependent advantages. On Electricity Transformer Temperature datasets, AIGS achieves preventative early-warning lead times of 8.31 (ETTm1) and 9.88 (ETTm2) steps under gradual degradation. On Performance Measurement System traffic datasets, it shows significantly faster post-shift recovery after abrupt mutations. On the highly noisy Weather dataset, it improves anomaly recall while resisting stochastic noise overfitting. These findings establish AIGS as a practical, plug-and-play adapter for resource-constrained edge monitoring systems. Codes will be made publicly available upon publication.
\end{abstract}





\maketitle

\section{Introduction}
Driven by the expansion of the WoT and urban computing, dynamic spatial-temporal streams in smart city edge sensing systems are rarely stationary. Traffic flow, weather variables, and industrial infrastructure metrics can drift at different speeds, a behavior commonly discussed as concept drift \cite{luLearningConceptDrift2018,gunasekaraRecurrentConceptDrifts2024}. In resource-constrained edge environments, the practical issue is whether the feature extractor can resolve the stability-plasticity dilemma: preserving an informative low-dimensional state during stable periods while rapidly adapting when the incoming stream changes, all without the massive retraining costs of heavy deep models. Online dimensionality reduction, including Incremental Principal Component Analysis (IPCA) \cite{wengCandidCovarianceFreeIncremental2003}, gives a standard starting point, but continuous operation still leaves this balance unresolved.

Existing methodologies often struggle to balance adaptation speed and computational overhead. Sliding-window approaches introduce severe computational redundancy with a time complexity of $\mathcal{O}\left(w\cdot d^2\right)$, where $w$ is the window size, and their reaction speed is rigidly tied to the window stride. Conversely, traditional online subspace tracking methods often employ fixed or monotonically decaying learning rates, which cause parameter ossification during sudden structural shifts. Furthermore, methods that adapt directly in the predictive space using end-to-end loss often face a severe reconstruction-decision gap. In low signal-to-noise ratio (SNR) streams, minimizing predictive error directly can easily overfit stochastic random-walk noise. Without an endogenous mechanism to filter feature-space mismatch before the prediction layer, heavy forecasters often extrapolate blindly during drift, and warning systems built on their outputs only react after downstream predictions have already failed.

To address these challenges under extreme computational constraints, a lightweight closed-loop state-aware adaptation framework called the Adaptive Incremental Gating System (AIGS) is proposed. Unlike traditional open-loop extractors or discrete drift detectors, AIGS treats representation learning itself as an endogenous sensing mechanism. Specifically, a normalized residual statistic, named the Shock Ratio, is introduced to dynamically compare the current reconstruction residual against recent local variation. This Endogenous Residual Feedback drives a Continuous Plasticity Controller that smoothly interpolates between learning plasticity and memory retention. The resulting module operates with strict $\mathcal{O}\left(k\cdot d\right)$ linear per-step computation, acting as a plug-and-play adapter that eliminates parameter ossification without suffering catastrophic forgetting.

The contributions of this work are summarized as follows:
\begin{itemize}
  \item \textbf{Endogenous Closed-loop Control.} We introduce the Shock Ratio, an Endogenous Residual Feedback mechanism that dynamically normalizes reconstruction error against a local moving average, bringing innovation-gating principles into unsupervised online representation learning.
  \item \textbf{Continuous Plasticity-Retention Interpolation.} A Continuous Plasticity Controller is designed via a Sigmoid gating mechanism. It continuously interpolates between learning plasticity (learning rate magnification) and memory retention (mean decay), entirely replacing traditional discrete hard resets and avoiding catastrophic forgetting.
  \item \textbf{State-Conditioned Dual-Signal Coupling.} AIGS produces a Joint State-Feature Representation, passing the manifold volatility state alongside the low-dimensional feature embedding to the downstream sequence model. This allows lightweight downstream temporal models to explicitly perceive structural shifts without structural resetting.
  \item \textbf{Empirical Efficiency \& Multi-Regime Validation.} Maintaining strict $\mathcal{O}\left(k\cdot d\right)$ time complexity and minimal memory overhead, AIGS systematically demonstrates Pareto optimality in edge resource allocation. Extensive validation across real-world streams reveals tailored domain-dependent advantages: preventative early warning under gradual degradation, rapid elastic recovery after abrupt mutations, and high anomaly recall in noisy environments.
\end{itemize}

\section{Related Work}
\subsection{Online Subspace Tracking and Robust PCA}
Classical online subspace tracking algorithms, including Oja’s rule with adaptive forgetting such as CCIPCA\cite{wengCandidCovarianceFreeIncremental2003}, OPAST\cite{yangProjectionApproximationSubspace1995} variants, and streaming PCA variants such as PETRELS \cite{chiPetrelsParallelSubspace2013}, GROUSE \cite{balzanoOnlineIdentificationTracking2010}, and GRASTA \cite{heOnlineRobustSubspace2011}, focus on matrix approximation and subspace estimation. They are useful when dealing with missing data or outliers. As feature extractors, however, their treatment of non-stationarity depends on predefined or externally injected heuristics, such as fixed exponential decay or static regularization. Recent combinations of online PCA and change detection can tune forgetting factors under concept drift\cite{xiaoOnlineRobustPrincipal2017}. However, these approaches remain predominantly open-loop estimators. They lack a closed-loop continuous plasticity control mechanism that leverages reconstruction states to dynamically regulate both the feature extraction step size and the downstream decisions simultaneously.

\subsection{Concept Drift-aware Adaptation}
The recent framework for drift-aware forecasting systems adapts in the output space through ensemble reweighting, such as ECW and GDW \cite{liuHandlingConceptDrift2023}, or in the model parameter space through continuous fine-tuning, as in Proceed \cite{zhaoProactiveModelAdaptation2025} and CEP \cite{zhanContinuousEvolutionPool2025}. Ensemble approaches require multiple base learners, increasing the computational cost of edge computing applications. ECW and GDW also use fixed base learners that may not fully capture evolving feature distributions. Additionally, parameter-space adaptation directly modifies deep sequence models using predictive loss, and this can make the model sensitive to short-term stochastic noise. The failure mode is close to the reconstruction-decision gap in low signal-to-noise ratio regimes. Proceed and CEP prioritize global updates, while FSNet \cite{phamLearningFastSlow2022}, OneNet \cite{zhangOneNetEnhancing2023}, DSOF \cite{lauFastSlowStreams2025}, D3A \cite{zhangAddressingConceptShift2024}, and DynaME \cite{hongDynamicMultiperiodExperts2026} emphasize online stream updating, dynamic assembly, detect-then-adapt behavior, incomplete future labels, or leakage-aware evaluation. Recent foundation and LLM-based forecasters extend cross-domain prediction through reprogramming, broad pretraining, or federated foundation modeling \cite{jinTimeLLMTimeSeries2024,goswamiMOMENTFamilyOpen2024,liuTimeFFMTowardsLMEmpowered2024}, while ablation evidence cautions that the LLM backbone itself may not always be the source of forecast gains \cite{tanAreLanguageModels2024}. These studies adapt primarily in the output space or the full model parameter space. In contrast, AIGS performs adaptation at the frontend feature representation layer, acting as a filter that resolves feature-space mismatch before sequence models process the data. This closed-loop frontend adjustment prevents the predictor from overfitting to stochastic noise during drift, especially under strict edge latency budgets.

\subsection{Statistical Drift Detection}
Classical change detectors address a complementary problem, where ADWIN \cite{bifetLearningTimechangingData2007}, DDM \cite{gamaLearningDriftDetection2004}, and CUSUM \cite{pageContinuousInspectionSchemes1954} are mostly discrete triggers applied to one-dimensional error streams. In forecast pipelines, they often initiate resets or retraining after a threshold violation. Recent work in time-series anomaly detection and prediction uses transformer association modeling, reconstruction-forecasting hybrids, contrastive representations, frequency patching, or forecast-based fine-tuning \cite{xuAnomalyTransformer2022,tuliTranADDeepTransformer2022,yangDCDetectorDual2023,wuCATCHChannelAware2025,zhouOnlineForecastingFineTuning2026}. Traditional detectors act as discrete boolean triggers that force model resets or retraining, often leading to catastrophic forgetting. AIGS fundamentally diverges by replacing discrete alarms with a continuous state-feature modulation, embedding the structural control signal directly within the feature extraction process.

\subsection{Distinct from Prior Endeavors}
Prior drift-aware subspace trackers can estimate an updated basis, but many rely on predefined forgetting or discrete statistical tests before memory changes. AIGS instead acts as a closed-loop controller around online representation learning. It uses the subspace tracking error, specifically the Shock Ratio, as endogenous feedback to regulate representation plasticity. Compared with output-space adaptation, the adjustment occurs before the sequence model receives the sample, which helps to separate feature-space mismatch from predictor error in noisy streams.

The proposed gating rule is also related to several foundational concept in adaptive filtering. For example, variable step-size LMS and NLMS \cite{kwongVariableStepSize1992} adjust convergence speed from error power, and Kalman filtering uses innovation-based gating for outlying measurements \cite{bar-shalomEstimationApplicationsTracking2001}. Similarly, adaptive forgetting in RLS or OPAST \cite{yangProjectionApproximationSubspace1995} discounts old data to improve tracking. AIGS applies this intuition at the representation level. Instead of relying on raw error power, it employs a dimensionless Shock Ratio to monitor  structural shifts in high-dimensional feature spaces.This design bridges the reconstruction-decision gap for downstream deep sequence models, allowing Oja, OPAST, or related operators to remain the core extractor while AIGS converts their residuals into a continuous downstream control signal. It operates as a closed-loop framework and replaces binary detection logic with a soft controller that interpolates between memory retention and plasticity to avoid catastrophic forgetting.

\section{Preliminaries and Problem Formulation}
This section establishes mathematical notations and formally defines the online representation learning and pre-warning tasks.

\subsection{Streaming Setting and Subspace}
Given an input stream $\mathcal{X}=\{x_1,\ldots,x_t,\ldots\}$ where $x_t\in\mathbb{R}^d$, the system tracks the dynamic mean vector $\mu_t$ to compute the centered vector $x_c$ = $x_t$ - $\mu_t$. The goal of online subspace learning is to maintain a low-dimensional weight matrix $W\in\mathbb{R}^{d\times k} (k < d)$ and produce a reconstruction error $L_t$ that indicates manifold variations~\cite{balzanoStreamingPCASubspace2018}.

\subsection{State Volatility Index}
In industrial IoT and meteorological streams, individual sensor readings are often dominated by high-frequency stochastic noise. To evaluate a system's sensitivity to structural manifold degradation rather than white-noise point forecasting, the State Volatility Index (SVI) is utilized. For ETT, anomaly pre-warning is evaluated through SVI, which captures the cross-sectional energy variance of the sensor topology. The SVI at time $t$ is formulated as the cross-sectional variance of sensor readings.
\begin{equation}
SVI_t=\frac{1}{N-1}\sum_{j=1}^{N}\left(x_{t,j}-\bar{x_t}\right)^2
\end{equation}
where $N$ denotes the number of features and $\bar{x_t}$ represents the cross-sectional mean. For the Weather dataset, the same formulation is used and the resulting target is referred to as the Weather Volatility Index (WVI), adopting established meteorological definitions\cite{benthModelingPricingFinancial2012}. For the PEMS traffic datasets, the evaluation focuses on traffic-flow prediction and post-shift recovery behavior rather than an SVI/WVI target.

\subsection{Problem Definition}
The online learner maps the centered stream and residual history to the next forecasting horizon.
\begin{equation}
\mathcal{F}\left(x_{c,\le t},L_{\le t}\right)\rightarrow\widehat{Y}_{t+1,\ldots,t+H}
\end{equation}
$Y$ denotes the predicted sequence over horizon $H$. For the ETT dataset, $Y$ represents the predicted SVI values; for the Weather dataset, it represents WVI values; and for the PEMS datasets, it refers to traffic-flow values. The model must minimize prediction error while adhering to a strict $O\left(k\cdot d\right)$ per-step computational complexity constraint. The formulation keeps the representation module manifold-aware while filtering steady-state stochastic noise.

\begin{figure*}
    \centering
  \includegraphics[width=\linewidth, trim=40 280 60 250, clip]{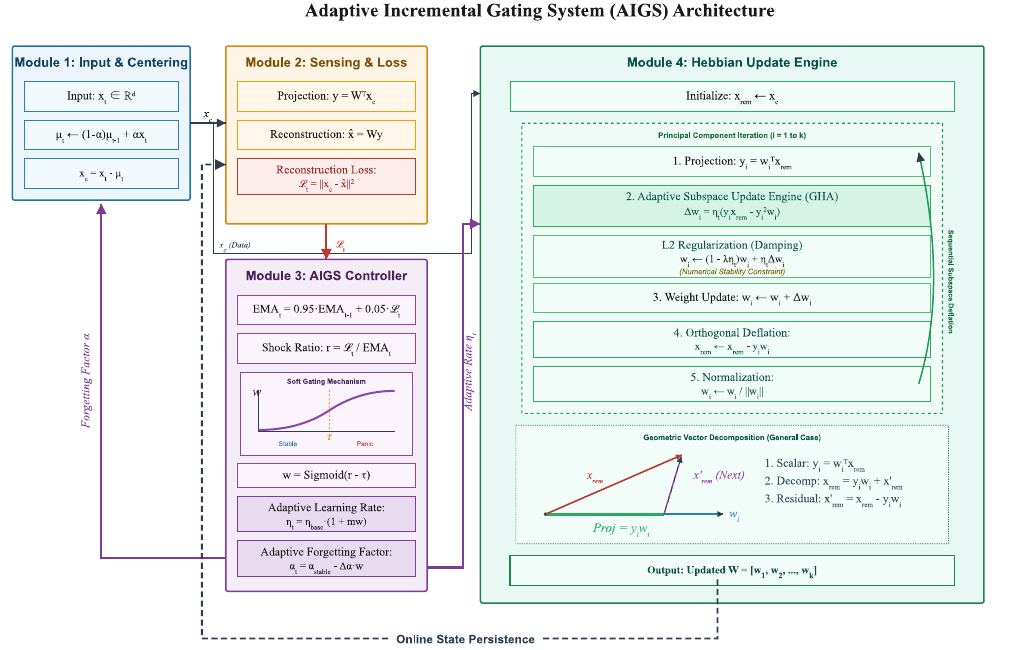}
  \caption{Adaptive Incremental Gating System (AIGS) Architecture}
  \Description{Architecture diagram showing the AIGS stream input, centering, subspace projection, residual sensing, Shock Ratio, soft gate, adaptive update, and downstream output.}
  \label{fig_aigs_arch}
\end{figure*}

\section{Proposed Model}
\subsection{Overall Architecture}
Figure~\ref{fig_aigs_arch} shows where the Adaptive Incremental Gating System (AIGS) places feedback within the feature-extraction pipeline. Unlike traditional open-loop feature extractors that suffer from parameter rigidity under concept drift, AIGS formulates online representation learning in streaming environments as a closed-loop sensing-control framework. The model consists of input centering and subspace projection, residual sensing and Shock Ratio calculation, the non-linear gating controller, and the adaptive subspace update engine. The intervention stays in the representation layer, reducing parameter rigidity under concept drift without replacing the downstream sequence model.

\subsection{Input Centering and Subspace Projection}
Given the input streaming vector $x_t\in\mathbb{R}^d$, the system maintains a dynamic mean vector $\mu_{t-1}\in\mathbb{R}^d$ to compute the centered observation $x_c=x_t-\mu_{t-1}$. To extract stable principal components, $x_c$ is projected onto a low-dimensional subspace defined by the weight matrix $W_{t-1} \in\mathbb{R}^{d\times k}$, where $k < d$. The encoding and decoding processes are formulated as follows.
\begin{equation}
y_t=W_{t-1}^Tx_c
\end{equation}
\begin{equation}
\widehat{x_t}=W_{t-1}y_t
\end{equation}
where $y_t\in\mathbb{R}^k$ represents the encoded low-dimensional feature embedding, and $\widehat{x_t}\in\mathbb{R}^d$ denotes the reconstructed data vector.

\subsection{Residual Sensing and Shock Ratio}
In non-stationary environments, traditional static tracking methods can show adaptation lag. To quantify the structural deviations induced by evolving data manifolds, the model monitors the real-time reconstruction error $\mathcal{L}_t\in\mathbb{R}$, defined as the Squared Error (SE) between the centered and reconstructed vectors.
\begin{equation}
\mathcal{L}_t=\left|\left|x_c-\widehat{x_t}\right|\right|^2
\end{equation}
Raw reconstruction error is unstable in these streams, especially when random-walk noise is high. AIGS therefore normalizes $\mathcal{L}_t$ against recent normal variation. The controller maintains an Exponential Moving Average (EMA) of the reconstruction loss.
\begin{equation}
EMA_t=0.95\cdot EMA_{t-1}+0.05\cdot\mathcal{L}_t
\end{equation}
The decay factor is set to 0.95, making the EMA a first-order temporal low-pass filter. At 0.95, high-frequency random noise in the reconstruction residual is attenuated and becomes the baseline for recent normal variation. The Shock Ratio is then calculated using the moving average as the denominator:
\begin{equation}
\begin{aligned}
r_t &\in\mathbb{R},\\
r_t &=\frac{\mathcal{L}_t}{EMA_t+\varepsilon}
\end{aligned}
\end{equation}
where $\varepsilon$ is a small constant used for numerical stability. Using absolute reconstruction error ($\mathcal{L}_t$) is insufficient because the background variance of the data stream fluctuates over time; a large absolute error may merely reflect increased environmental noise rather than true concept drift. To address this, AIGS introduces the Shock Ratio ($r_t$), a self-standardizing, dimensionless metric that normalizes $\mathcal{L}_t$ against recent normal variation. High values of $r_t$ mark residual excursions relative to the local EMA baseline, effectively acting as a high-pass filter that isolates sudden structural drifts from low-frequency background noise.

\subsection{Non-linear Gating Controller}
Traditional streaming change detectors typically employ discrete hard triggers. Hard triggers are useful when an alert is needed, but their binary reset action is too coarse for feature extraction and may discard useful historical states. AIGS maps the residual ratio to a non-linear soft gate, allowing  update to occur gradually between steady-state periods and structural transition. The adaptive gating weight $w\in\left(0,1\right)$ is derived via a shifted Sigmoid function:
\begin{equation}
w=\frac{1}{1+e^{-c\left(r_t-\tau\right)}}
\end{equation}
where $\tau$ denotes the activation threshold. The steepness factor $c$ can be varied; reported experiments use $c=1$ to maintain a standard sigmoid transition and unit-scale responsiveness. Using $c=1$ also keeps the closed-loop feedback test separate from domain-specific parameter sensitivity. The gate modulates learning plasticity through a magnified step size \cite{duchiAdaptiveSubgradientMethods2011}, giving the model a way to recalibrate during structural shocks without the hard loss of historical state associated with discrete resets. The adaptive learning rate $\eta_t$ and the memory retention coefficient $\alpha_t$ are updated using linear scaling:
\begin{equation}
\begin{aligned}
\eta_t &= \eta_{base}\cdot\left(1+M\cdot w\right),\\
\alpha_t &= \alpha_{stable}-\Delta\alpha\cdot w
\end{aligned}
\end{equation}
where $\eta_{base}$ is the base step size, $M$ is the magnification factor, $\alpha_{stable}$ is the baseline retention coefficient, and $\Delta \alpha$ defines the adjustment margin. The derived $\alpha_t$ loops back into the recursive mean update $\mu_t = \alpha_t\mu_{t-1} + (1 - \alpha_t)x_t$. A larger $\alpha_t$ will make the center close to historical state, while a smaller $\alpha_t$ will give new observation higher weight and shortens the effective memory time. During a structural shift, a larger $w$ increase the learning rate while reducing outdated statistical state through a shorter observation window. Passing $w$ downstream allows the addition of a state metric that sense the volatility of feature-space, thus providing the sequence model with an endogenous context of the manifold.

\subsection{Adaptive Subspace Update Engine}
To continuously track the evolving subspace, the subspace update uses a modified Generalized Hebbian Algorithm (GHA) \cite{sangerOptimalUnsupervisedLearning1989}. To encourage orthogonality of the extracted principal components, the i-th principal component is computed iteratively from the residual left after earlier components have been removed:
\begin{equation}
\begin{aligned}
x_{rem,i} &\in\mathbb{R}^d,\\
x_{rem,i} &=x_c-\sum_{j=1}^{i-1}{y_jW_j}
\end{aligned}
\end{equation}
where $W_j\in\mathbb{R}^d$ denotes the $j$-th column vector of the weight matrix. The weight vector is then updated with the adaptive learning rate $\eta_t$ and an $L_2$-norm damping term:
\begin{equation}
W_i^{new}=\left(1-\eta_t\lambda\right)W_i^{old}+\eta_t\left(y_ix_{rem,i}-y_i^2W_i^{old}\right)
\end{equation}
where $\lambda$ specifies the weight decay penalty. The factor $(1-\eta_t\lambda)$ restricts the spectral growth of the weight matrix and reduces the risk of gradient explosion when $\eta_t$ increases. In practice, the sequential deflation step isolates residual variance to maintain structural integrity. The update follows the component-separation principle of GHA and is consistent with established stochastic approximation limits.

\subsection{Convergence and Tracking Stability Analysis}
In continuous data streams, AIGS navigates the inherent tension between step size, noise variance, and tracking bias \cite{kushnerStochasticApproximationRecursive2003}. The tracking error typically follows the $O(\delta/(\eta_{t}\mu))$ scaling derived from stochastic approximation limits \cite{balzanoStreamingPCASubspace2018,kushnerStochasticApproximationRecursive2003}, where $\delta$ represents the magnitude of the concept drift and $\mu$ denotes the local manifold curvature. While fixed step-size methods suffer from error expansion or slow correction during abrupt regime shifts, the gating mechanism in AIGS raises the effective step size through $M$ and $w$ when the residual ratio rises. This dynamic adjustment theoretically compresses the drift-induced tracking error scale to $O(\delta/(\eta_{base}(1+M)\mu))$ \cite{sangerOptimalUnsupervisedLearning1989}, effectively reducing tracking bias. System stability is further characterized by bias-variance trade-offs under non-stationary conditions. The adaptive increase in $\eta_t$ improves tracking during structural shifts, while the EMA decay and $\alpha_t$ act as low-pass filters to suppress noise-induced variance. This mechanism ensures that the total representation error remains bounded even in high-noise environments. Since the architecture avoids the calculation of complete covariance matrices, the time complexity remains $O(k\cdot d)$ per algorithmic step, strictly meeting the latency constraints of real-time industrial monitoring.

\subsection{Joint State-Feature Representation and Downstream Integration}
AIGS is placed before the sequence model and outputs a reconstruction augmented embedding. At time $t$, the downstream vector is $v_{t} = [z_{t}, y_{t}, \phi_{t}]$, where $z_t$ stores the principal spatial components and $\phi_t$ stores the gated residual state. Because $\phi_t$ comes from the gated reconstruction residual, structural volatility becomes part of the temporal input rather than a separate alarm. Traditional predictors experience an inherent time lag, as they must accumulate predictive loss to sense drift. In contrast, AIGS supplies a Joint State-Feature Representation. The downstream GRU receives both the compact feature representation and a state channel explicitly indicating whether the underlying feature manifold is stable or shifting. This allows the lightweight temporal learner to immediately perceive structural shifts during the forward pass, adjusting its internal state-tracking responsiveness without requiring structural resets.

\begin{table}[htbp]
  \centering
  \caption{Summary of datasets}
  \label{tab_dataset_stats}
  \begin{tabular}{lccc}
    \toprule
    \textbf{Datasets} & \textbf{Dimension} & \textbf{Time Interval} & \textbf{Timesteps} \\
    \midrule
    PEMS04       & 307  & 5 mins  & 16,992 \\
    PEMS08       & 170  & 5 mins  & 17,856 \\
    ETTm1        & 7    & 15 mins & 69,680 \\
    ETTm2        & 7    & 15 mins & 69,680 \\
    Weather      & 14   & 10 mins & 52,259 \\
    \bottomrule
  \end{tabular}
\end{table}

\section{Experimental Setup}
\subsection{Datasets and Forecasting Tasks}
The experiment covers real-world datasets from three non-stationary domains, as shown in Table~\ref{tab_dataset_stats}. The industrial IoT setting uses ETTm1 and ETTm2 \cite{zhouInformerEfficientTransformer2021}, with the State Volatility Index (SVI) predicted at horizons {8, 16, 32, 48, 96, 192, 288, 672, 1344}. The traffic setting uses PEMS04 and PEMS08 \cite{guoAttentionBasedSpatialtemporal2019} for flow recovery behavior at horizons {3, 6, 12, 24}. The meteorology part uses the Jena Climate 2016 dataset \cite{maxPlanckJenaWeatherData2026}, where the Weather Volatility Index (WVI) is tested at horizons {36, 72, 144, 288, 432, 1008, 1440}.

Predicting these volatility indices allows the model to identify transitions from stable regimes to unstable states. The target choice keeps the evaluation focused on signal energy distribution rather than stochastic white noise, which is where the reconstruction-decision gap matters. Following prior online forecasting protocols, each stream is split into warm-up and evaluation subsets with a one-to-one ratio. AIGS and all baselines use the same split, arrival order, objective function, and online protocol.

\subsection{Baseline Methods}
AIGS+GRU is benchmarked against methods that cover lightweight forecasting, attention-based forecasting, adaptive sequence modeling, streaming adaptation, and traffic graph modeling. DLinear \cite{zengAreTransformersEffective2023} represents a lightweight forecasting baseline. PatchTST \cite{nieTimeSeriesWorth2022} and iTransformer \cite{liuItransformerInvertedTransformers2024} represent attention-based long-horizon forecasters. Naked GRU and ReVIN+GRU \cite{kimReversibleInstanceNormalization2021} serve as the primary adaptive sequence baselines. CEP, FSNet, and ADWIN+\allowbreak{}GRU represent streaming methods. T-GCN \cite{zhaoTGCNTemporalGraph2020} represents STGNN architecture adaptations for traffic data.
All baseline models are run under the same online prediction protocol with a unified objective function so that differences are evaluated under shared streaming conditions. To verify extractor-agnostic robustness, AIGS is also evaluated across alternative feature extractors.

\subsection{Ensemble Strategy and Implementation Details}
A three-model ensemble is used for all learning models following deep ensemble practice \cite{lakshminarayananSimpleScalablePredictive2017}. Each member has the same architecture but a different initialization derived from an incremented base seed. The members read the stream synchronously through one pending queue and optimize independently. During drift, AIGS modulates the learning rates for all members. Prediction uses mean aggregation across the parallel outputs to reduce variance and improve stability against transient volatility. Trial runs on ETT with base seeds 42, 999, and 2026 showed the same qualitative trend, so the reported experiments fix the global base seed at 42 for reproducibility. 

\subsection{Implementation Details}
The AIGS framework utilizes a three-model ensemble to ensure predictive stability across non-stationary intervals. The input feature subspace has 3 principal components by default. For the higher-dimensional PEMS datasets, it is increased to 16; for Weather, it is increased to 6. The extracted features pass to a GRU with a 64-unit hidden state. Online optimization uses a base learning rate of 0.001 and a buffer size of 16 samples. The default switch threshold is 0.5 and the default magnification factor is 1. For Weather, these settings are 1.0 and 5, respectively, as the reported domain-specific configuration. The stable forgetting factor is 0.95 and the volatility factor is 0.8, corresponding to effective observation windows of about 20 and 5 time steps. All experiments are run on a single \textbf{NVIDIA GeForce RTX 5070 Ti Laptop GPU}. Generative AI was used to assist in writing minor code snippets, while all core algorithms and final implementations were independently developed and verified by the authors.

\subsection{Evaluation Metrics}
To quantitatively assess proactive sensing capabilities, it is measured with Average Pre-warning Lead Time ($AvgLead$). The ground-truth regime-shift events, denoted as $A=\{t_{start,i},t_{end,i}\}$, are extracted from the raw target sequence with a rolling 3-sigma rule. An event is counted only when the signal remains beyond three standard deviations from the rolling mean for at least two consecutive time steps. The two-step rule avoids treating a single noisy point as an event and keeps event extraction independent from model predictions. The predictive alarm uses the model output signal with an attenuation factor of 0.85 and is triggered when the signal exceeds 0.6 for more than three consecutive steps. These event extraction thresholds strictly follow established prior configurations\cite{tuliTranADDeepTransformer2022, montgomeryIntroductionStatisticalQuality2020}, ensuring benchmarking fairness without requiring additional robustness tests. The $AvgLead$ metric is formulated as follows.
\begin{equation}
AvgLead=\frac{1}{\left|H\right|}\sum_{i\in H}\left(t_{start,i}-T_{trigger,i}\right)
\end{equation}
where $H$ denotes the set of events that trigger an alert. $AvgLead$ is the mean offset between the event start and the first warning trigger. Since it is computed only over triggered events, it must be read together with recall and false alarm rate (FAR). Lead time alone does not penalize missed detections. This consideration is particularly important for PEMS because traffic streams often exhibit sudden non-stationary mutations rather than gradual cumulative errors.

To evaluate post-mutation recovery capability in
these scenarios, Anomaly Recovery Time ($ART$) is introduced. Let $e_t=\left|y_t-\widehat{y_t}\right|$ represent the absolute error, with $\mu_t$ and $\sigma_t$ denoting its rolling mean and standard deviation over a window $W$=30. An error spike is identified at time $t_s$ if $e_{t_s}>\mu_{t_s}+3\sigma_{t_s}$. The recovery duration $T$ is the number of later steps required for the error to fall below the pre-spike threshold $\mu_{t_s-1}+\sigma_{t_s-1}$. The overall ART is calculated as the average recovery duration
across identified spikes, so lower values indicate faster recovery after abrupt distribution shifts.

\begin{table*}[htbp]
\centering
\caption{PEMS04 and PEMS08 Experimental Results}
\label{tab_pems_results}
\resizebox{\textwidth}{!}{
\begin{tabular}{lcccccccccccc}
\toprule
 & \multicolumn{6}{c}{PEMS04} & \multicolumn{6}{c}{PEMS08} \\
\cmidrule(lr){2-7} \cmidrule(lr){8-13}
 & \multicolumn{3}{c}{Short-term (3 6)} & \multicolumn{3}{c}{Long-term (12 24)} & \multicolumn{3}{c}{Short-term (3 6)} & \multicolumn{3}{c}{Long-term (12 24)} \\
\cmidrule(lr){2-4} \cmidrule(lr){5-7} \cmidrule(lr){8-10} \cmidrule(lr){11-13}
Model & RMSE & MAE & ART & RMSE & MAE & ART & RMSE & MAE & ART & RMSE & MAE & ART \\
\midrule
AIGS+GRU & $\underline{30.56\pm 0.07}$ & $\underline{\underline{18.86\pm 0.50}}$ & $\mathbf{12.76\pm 1.44}$ & $\mathbf{34.35\pm 3.90}$ & $\mathbf{23.38\pm 3.93}$ & $\mathbf{17.38\pm 2.44}$ & $\underline{28.65\pm 0.26}$ & $\underline{\underline{18.47\pm 0.33}}$ & $\mathbf{10.96\pm 3.28}$ & $\mathbf{34.98\pm 6.39}$ & $\mathbf{24.81\pm 5.97}$ & $\mathbf{16.42\pm 0.59}$ \\
T-GCN & $48.29\pm 2.04$ & $36.68\pm 1.76$ & $28.40\pm 3.39$ & $63.09\pm 9.30$ & $49.20\pm 7.45$ & $30.96\pm 3.98$ & $47.83\pm 1.79$ & $36.38\pm 1.23$ & $32.37\pm 0.10$ & $59.25\pm 7.55$ & $45.42\pm 6.01$ & $27.80\pm 6.44$ \\
Naked\_GRU & $\underline{\underline{27.49\pm 1.04}}$ & $\underline{22.06\pm 1.02}$ & $\underline{19.56\pm 0.23}$ & $52.36\pm 15.33$ & $41.40\pm 11.55$ & $30.41\pm 7.24$ & $\underline{\underline{26.76\pm 3.23}}$ & $\underline{19.46\pm 2.52}$ & $17.91\pm 0.88$ & $43.46\pm 11.40$ & $32.73\pm 9.50$ & $31.07\pm 6.20$ \\
iTransformer & $32.75\pm 2.29$ & $23.88\pm 1.88$ & $30.33\pm 6.36$ & $\underline{48.89\pm 10.72}$ & $\underline{36.96\pm 8.50}$ & $33.90\pm 0.80$ & $29.01\pm 2.02$ & $20.91\pm 1.66$ & $22.44\pm 4.37$ & $\underline{43.18\pm 9.23}$ & $\underline{32.64\pm 7.38}$ & $27.22\pm 2.38$ \\
PatchTST & $67.63\pm 3.76$ & $47.76\pm 3.51$ & $34.15\pm 2.38$ & $76.72\pm 0.03$ & $54.18\pm 1.33$ & $41.81\pm 1.76$ & $53.32\pm 0.26$ & $39.51\pm 0.26$ & $34.22\pm 2.06$ & $53.92\pm 0.32$ & $40.37\pm 0.05$ & $34.89\pm 0.01$ \\
DLinear & $34.78\pm 2.79$ & $24.88\pm 2.99$ & $22.61\pm 0.31$ & $53.99\pm 12.05$ & $43.94\pm 11.46$ & $38.21\pm 10.08$ & $31.93\pm 2.44$ & $22.89\pm 2.63$ & $15.74\pm 4.34$ & $48.73\pm 10.45$ & $39.63\pm 10.03$ & $38.98\pm 4.03$ \\
ReVIN+GRU & $\mathbf{21.35\pm 3.24}$ & $\mathbf{14.50\pm 2.77}$ & $\underline{\underline{16.43\pm 0.43}}$ & $\underline{\underline{46.91\pm 18.63}}$ & $\underline{\underline{35.79\pm 14.75}}$ & $\underline{\underline{25.90\pm 1.39}}$ & $\mathbf{16.72\pm 2.12}$ & $\mathbf{12.14\pm 1.77}$ & $\underline{\underline{13.47\pm 1.04}}$ & $\underline{\underline{41.53\pm 15.93}}$ & $\underline{\underline{31.62\pm 12.80}}$ & $\underline{\underline{24.70\pm 1.55}}$ \\
CEP & $36.85\pm 0.83$ & $28.55\pm 0.47$ & $23.72\pm 6.26$ & $53.83\pm 9.76$ & $42.56\pm 7.41$ & $36.33\pm 3.90$ & $31.53\pm 1.84$ & $23.93\pm 0.71$ & $\underline{14.12\pm 0.53}$ & $48.92\pm 14.12$ & $35.36\pm 10.43$ & $34.26\pm 8.01$ \\
FSNet & $57.68\pm 1.24$ & $41.47\pm 1.88$ & $26.02\pm 1.82$ & $64.58\pm 10.06$ & $48.79\pm 8.81$ & $37.63\pm 7.02$ & $32.19\pm 2.63$ & $23.89\pm 3.85$ & $16.51\pm 0.50$ & $52.77\pm 11.62$ & $41.11\pm 9.28$ & $35.75\pm 6.06$ \\
ADWIN+GRU & $44.14\pm 1.19$ & $33.71\pm 0.36$ & $25.85\pm 0.04$ & $83.64\pm 19.04$ & $64.74\pm 15.22$ & $\underline{26.60\pm 10.98}$ & $49.90\pm 16.84$ & $37.58\pm 13.10$ & $22.44\pm 1.83$ & $86.85\pm 13.55$ & $69.75\pm 13.75$ & $\underline{26.02\pm 0.33}$ \\
\bottomrule
\end{tabular}
}
\end{table*}

\begin{table*}[htbp]
\centering
\caption{Experimental Results on ETTm1 Dataset}
\label{tab_ettm1_results}
\resizebox{\textwidth}{!}{
\begin{tabular}{lccccccccc}
\toprule
\multirow{2}{*}{Model} & \multicolumn{3}{c}{Short-term (8 16 32)} & \multicolumn{3}{c}{Mid-term (48 96 192)} & \multicolumn{3}{c}{Long-term (288 672 1344)} \\
\cmidrule(lr){2-4} \cmidrule(lr){5-7} \cmidrule(lr){8-10}
 & RMSE & MAE & Avg\_Lead & RMSE & MAE & Avg\_Lead & RMSE & MAE & Avg\_Lead \\
\midrule
AIGS+GRU & $\underline{\underline{1.49\pm 0.15}}$ & $\underline{\underline{1.06\pm 0.10}}$ & $\mathbf{8.31\pm 2.95}$ & $\underline{1.60\pm 0.08}$ & $\underline{\underline{1.12\pm 0.09}}$ & $\mathbf{8.09\pm 0.92}$ & $\mathbf{1.65\pm 0.01}$ & $\mathbf{1.18\pm 0.02}$ & $\mathbf{7.14\pm 0.87}$ \\
ADWIN+GRU & $1.66\pm 0.04$ & $1.20\pm 0.03$ & $-2.55\pm 0.39$ & $1.67\pm 0.03$ & $1.19\pm 0.02$ & $-3.74\pm 0.94$ & $1.70\pm 0.01$ & $\underline{1.22\pm 0.01}$ & $-0.64\pm 0.91$ \\
CEP & $1.66\pm 0.03$ & $1.17\pm 0.05$ & $-3.71\pm 2.48$ & $1.62\pm 0.08$ & $\underline{1.13\pm 0.08}$ & $-3.56\pm 1.02$ & $\underline{1.68\pm 0.01}$ & $\underline{\underline{1.19\pm 0.02}}$ & $-1.00\pm 1.26$ \\
DLinear & $1.58\pm 0.10$ & $1.15\pm 0.09$ & $-3.29\pm 1.18$ & $\mathbf{1.55\pm 0.07}$ & $\mathbf{1.08\pm 0.06}$ & $-3.90\pm 1.28$ & $1.70\pm 0.01$ & $1.24\pm 0.02$ & $\underline{0.51\pm 5.70}$ \\
FSNet & $1.56\pm 0.13$ & $1.12\pm 0.10$ & $-2.23\pm 1.41$ & $1.62\pm 0.07$ & $1.14\pm 0.08$ & $-4.51\pm 0.49$ & $\underline{1.68\pm 0.01}$ & $\underline{\underline{1.19\pm 0.02}}$ & $-1.33\pm 1.73$ \\
T-GCN & $1.70\pm 0.01$ & $1.24\pm 0.02$ & $\underline{0.32\pm 0.52}$ & $1.71\pm 0.01$ & $1.23\pm 0.03$ & $\underline{-1.87\pm 0.93}$ & $1.72\pm 0.00$ & $1.24\pm 0.01$ & $-0.58\pm 1.26$ \\
Naked\_GRU & $\underline{1.53\pm 0.14}$ & $\underline{1.10\pm 0.10}$ & $-1.49\pm 2.65$ & $1.61\pm 0.08$ & $\underline{1.13\pm 0.08}$ & $-4.34\pm 0.82$ & $\underline{\underline{1.67\pm 0.01}}$ & $\underline{\underline{1.19\pm 0.02}}$ & $-1.02\pm 1.31$ \\
PatchTST & $1.92\pm 0.05$ & $1.45\pm 0.03$ & $-2.18\pm 0.59$ & $1.95\pm 0.03$ & $1.48\pm 0.03$ & $\underline{\underline{-0.69\pm 0.93}}$ & $2.03\pm 0.06$ & $1.54\pm 0.03$ & $\underline{\underline{1.68\pm 2.35}}$ \\
ReVIN+GRU & $1.61\pm 0.15$ & $1.17\pm 0.13$ & $\underline{\underline{0.69\pm 1.10}}$ & $1.71\pm 0.12$ & $1.21\pm 0.15$ & $-3.07\pm 2.66$ & $1.81\pm 0.02$ & $1.28\pm 0.04$ & $-4.54\pm 1.60$ \\
iTransformer & $\mathbf{1.44\pm 0.08}$ & $\mathbf{0.96\pm 0.06}$ & $-4.79\pm 0.95$ & $\underline{\underline{1.58\pm 0.06}}$ & $\mathbf{1.08\pm 0.05}$ & $-3.15\pm 0.93$ & $1.71\pm 0.02$ & $1.23\pm 0.03$ & $-2.72\pm 0.85$ \\
\bottomrule
\end{tabular}
}
\end{table*}

\begin{table*}[htbp]
\centering
\caption{Experimental Results on ETTm2 Dataset}
\label{tab_ettm2_results}
\resizebox{\textwidth}{!}{
\begin{tabular}{lccccccccc}
\toprule
\multirow{2}{*}{Model} & \multicolumn{3}{c}{Short-term (8 16 32)} & \multicolumn{3}{c}{Mid-term (48 96 192)} & \multicolumn{3}{c}{Long-term (288 672 1344)} \\
\cmidrule(lr){2-4} \cmidrule(lr){5-7} \cmidrule(lr){8-10}
 & RMSE & MAE & Avg\_Lead & RMSE & MAE & Avg\_Lead & RMSE & MAE & Avg\_Lead \\
\midrule
AIGS+GRU & $\underline{1.90\pm 0.30}$ & $\underline{1.45\pm 0.24}$ & $\underline{\underline{5.87\pm 1.15}}$ & $\mathbf{2.21\pm 0.11}$ & $\underline{\underline{1.65\pm 0.10}}$ & $\mathbf{9.88\pm 0.96}$ & $\mathbf{2.28\pm 0.02}$ & $\mathbf{1.70\pm 0.02}$ & $\mathbf{8.22\pm 0.86}$ \\
ADWIN+GRU & $1.91\pm 0.32$ & $1.48\pm 0.26$ & $4.09\pm 2.68$ & $2.31\pm 0.08$ & $1.73\pm 0.08$ & $0.08\pm 1.21$ & $\underline{2.34\pm 0.04}$ & $\underline{1.76\pm 0.03}$ & $1.68\pm 3.53$ \\
CEP & $1.91\pm 0.32$ & $1.47\pm 0.25$ & $3.97\pm 2.22$ & $\underline{2.26\pm 0.11}$ & $\underline{1.70\pm 0.11}$ & $-0.27\pm 1.55$ & $\underline{2.34\pm 0.04}$ & $\underline{1.76\pm 0.03}$ & $1.62\pm 3.45$ \\
DLinear & $\underline{\underline{1.82\pm 0.21}}$ & $\underline{\underline{1.37\pm 0.15}}$ & $5.43\pm 2.65$ & $\underline{\underline{2.23\pm 0.12}}$ & $\mathbf{1.63\pm 0.10}$ & $0.14\pm 1.65$ & $2.63\pm 0.10$ & $1.93\pm 0.07$ & $0.69\pm 1.89$ \\
FSNet & $1.92\pm 0.31$ & $1.47\pm 0.24$ & $3.10\pm 1.19$ & $\underline{2.27\pm 0.07}$ & $\underline{1.70\pm 0.07}$ & $2.21\pm 0.82$ & $\underline{\underline{2.30\pm 0.03}}$ & $\underline{\underline{1.73\pm 0.02}}$ & $1.95\pm 3.67$ \\
T-GCN & $2.31\pm 0.18$ & $1.78\pm 0.13$ & $\underline{5.73\pm 0.34}$ & $2.43\pm 0.09$ & $1.85\pm 0.07$ & $3.78\pm 1.52$ & $2.50\pm 0.01$ & $1.90\pm 0.00$ & $2.35\pm 1.00$ \\
Naked\_GRU & $1.91\pm 0.32$ & $1.48\pm 0.26$ & $4.09\pm 2.68$ & $2.28\pm 0.12$ & $1.71\pm 0.11$ & $0.33\pm 1.27$ & $2.35\pm 0.04$ & $1.77\pm 0.03$ & $1.84\pm 3.52$ \\
PatchTST & $3.84\pm 0.05$ & $2.80\pm 0.03$ & $\mathbf{8.93\pm 0.69}$ & $4.68\pm 0.67$ & $3.45\pm 0.46$ & $\underline{\underline{6.42\pm 2.25}}$ & $4.62\pm 0.35$ & $3.38\pm 0.23$ & $\underline{\underline{6.20\pm 5.33}}$ \\
ReVIN+GRU & $2.03\pm 0.35$ & $1.53\pm 0.28$ & $2.37\pm 1.59$ & $2.53\pm 0.18$ & $1.79\pm 0.15$ & $\underline{5.43\pm 1.80}$ & $3.11\pm 0.11$ & $2.26\pm 0.12$ & $2.07\pm 3.93$ \\
iTransformer & $\mathbf{1.79\pm 0.17}$ & $\mathbf{1.29\pm 0.12}$ & $-0.37\pm 0.84$ & $2.38\pm 0.22$ & $\underline{1.70\pm 0.17}$ & $1.22\pm 0.31$ & $2.92\pm 0.09$ & $2.18\pm 0.10$ & $\underline{3.38\pm 2.91}$ \\
\bottomrule
\end{tabular}
}
\end{table*}

\begin{table*}[htbp]
\centering
\caption{Experimental Results on Weather Dataset}
\label{tab_weather_results}
\resizebox{\textwidth}{!}{
\begin{tabular}{lcccccc}
\toprule
\multirow{2}{*}{Model} & \multicolumn{2}{c}{Short-term (36 72)} & \multicolumn{2}{c}{Mid-term (144 288 432)} & \multicolumn{2}{c}{Long-term (1008 1440)} \\
\cmidrule(lr){2-3} \cmidrule(lr){4-5} \cmidrule(lr){6-7}
 & RMSE & MAE & RMSE & MAE & RMSE & MAE \\
\midrule
AIGS+GRU & $\underline{0.27\pm 0.01}$ & $0.21\pm 0.01$ & $\underline{0.27\pm 0.00}$ & $\underline{0.21\pm 0.00}$ & $\underline{\underline{0.27\pm 0.00}}$ & $\underline{\underline{0.21\pm 0.00}}$ \\
ADWIN+GRU & $\underline{0.27\pm 0.00}$ & $\underline{0.20\pm 0.00}$ & $\mathbf{0.25\pm 0.01}$ & $\mathbf{0.19\pm 0.00}$ & $\mathbf{0.26\pm 0.01}$ & $\mathbf{0.20\pm 0.00}$ \\
CEP & $\underline{0.27\pm 0.01}$ & $\underline{0.20\pm 0.01}$ & $\underline{\underline{0.26\pm 0.01}}$ & $\underline{\underline{0.20\pm 0.01}}$ & $\underline{0.28\pm 0.01}$ & $\underline{0.22\pm 0.01}$ \\
DLinear & $\underline{\underline{0.26\pm 0.00}}$ & $\underline{\underline{0.19\pm 0.00}}$ & $0.29\pm 0.02$ & $0.22\pm 0.02$ & $0.32\pm 0.00$ & $0.25\pm 0.00$ \\
FSNet & $\underline{\underline{0.26\pm 0.00}}$ & $0.21\pm 0.01$ & $\underline{\underline{0.26\pm 0.01}}$ & $\underline{\underline{0.20\pm 0.01}}$ & $\underline{\underline{0.27\pm 0.00}}$ & $\underline{\underline{0.21\pm 0.00}}$ \\
T-GCN & $0.32\pm 0.01$ & $0.25\pm 0.01$ & $0.31\pm 0.01$ & $0.24\pm 0.00$ & $0.32\pm 0.00$ & $0.25\pm 0.00$ \\
Naked\_GRU & $\underline{\underline{0.26\pm 0.01}}$ & $\underline{0.20\pm 0.01}$ & $\underline{0.27\pm 0.01}$ & $\underline{0.21\pm 0.01}$ & $0.29\pm 0.00$ & $\underline{0.22\pm 0.00}$ \\
PatchTST & $0.43\pm 0.03$ & $0.33\pm 0.02$ & $0.46\pm 0.01$ & $0.35\pm 0.01$ & $0.46\pm 0.06$ & $0.35\pm 0.04$ \\
ReVIN+GRU & $0.30\pm 0.03$ & $0.22\pm 0.02$ & $0.31\pm 0.04$ & $0.24\pm 0.03$ & $0.40\pm 0.00$ & $0.31\pm 0.00$ \\
iTransformer & $\mathbf{0.24\pm 0.01}$ & $\mathbf{0.18\pm 0.01}$ & $0.33\pm 0.03$ & $0.25\pm 0.02$ & $0.40\pm 0.01$ & $0.31\pm 0.00$ \\
\bottomrule
\end{tabular}
}
\end{table*}

\begin{figure}[tbp]
    \centering
    \includegraphics[width=0.9\linewidth]{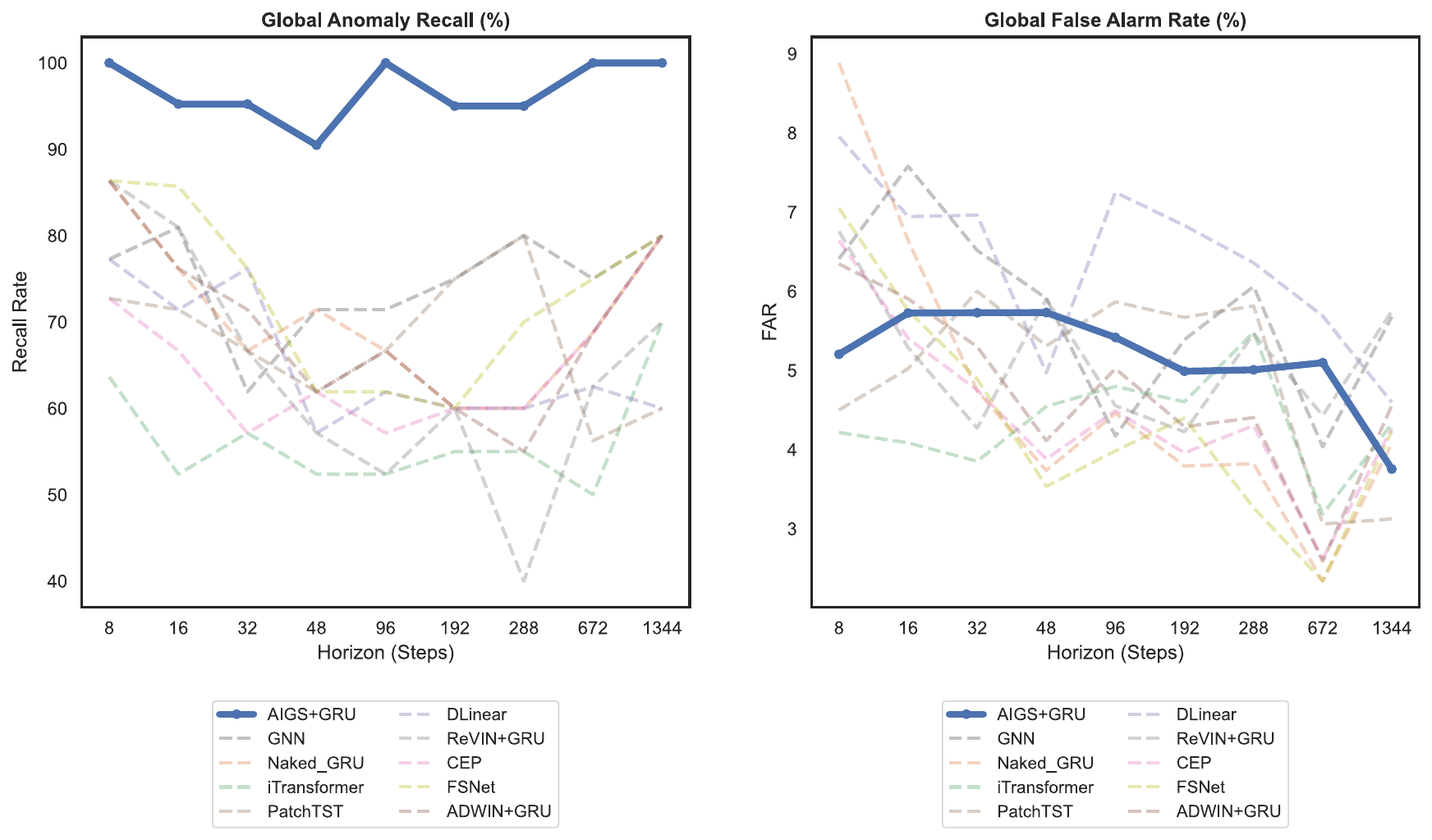}
    \caption{Global Anomaly Recall and False Alarm Rate (FAR) Across Different Prediction Horizons.}
    \Description{Plot comparing anomaly recall and false alarm rate across prediction horizons under the rolling 3-sigma event protocol.}
    \label{fig_warning_efficiency}
\end{figure}

\begin{figure}[tbp]
    \centering
    \includegraphics[trim=0cm 0cm 0cm 1cm, clip,width=0.9\linewidth]{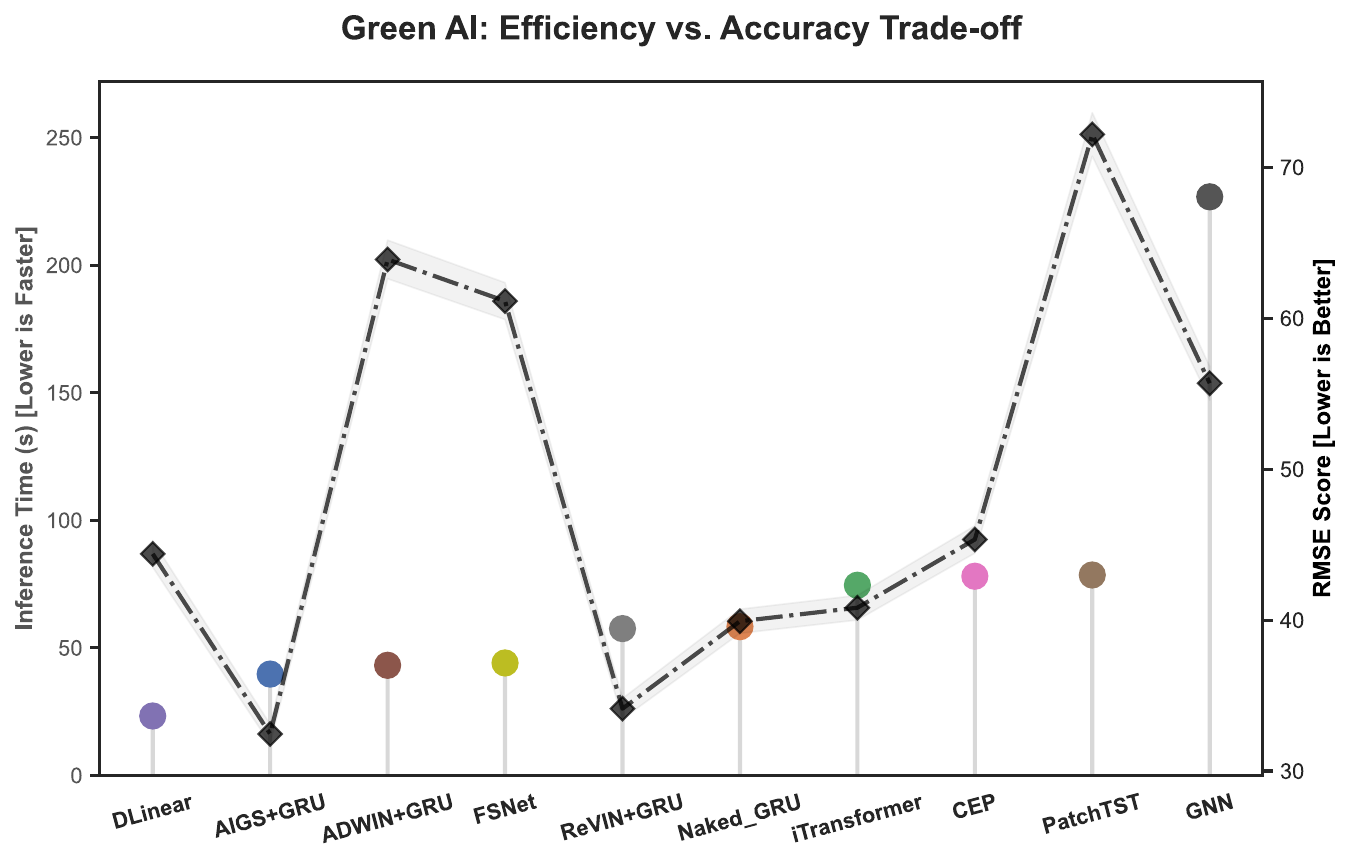}
\caption{Analysis of the Trade-off between Inference Time (s) and RMSE Score.}
    \Description{Lollipop plot comparing inference time and RMSE for evaluated models to show the efficiency and accuracy trade-off.}
    \label{fig_efficiency_accuracy}
\end{figure}

\begin{figure}[tbp]
    \centering
    \includegraphics[width=0.9\linewidth]{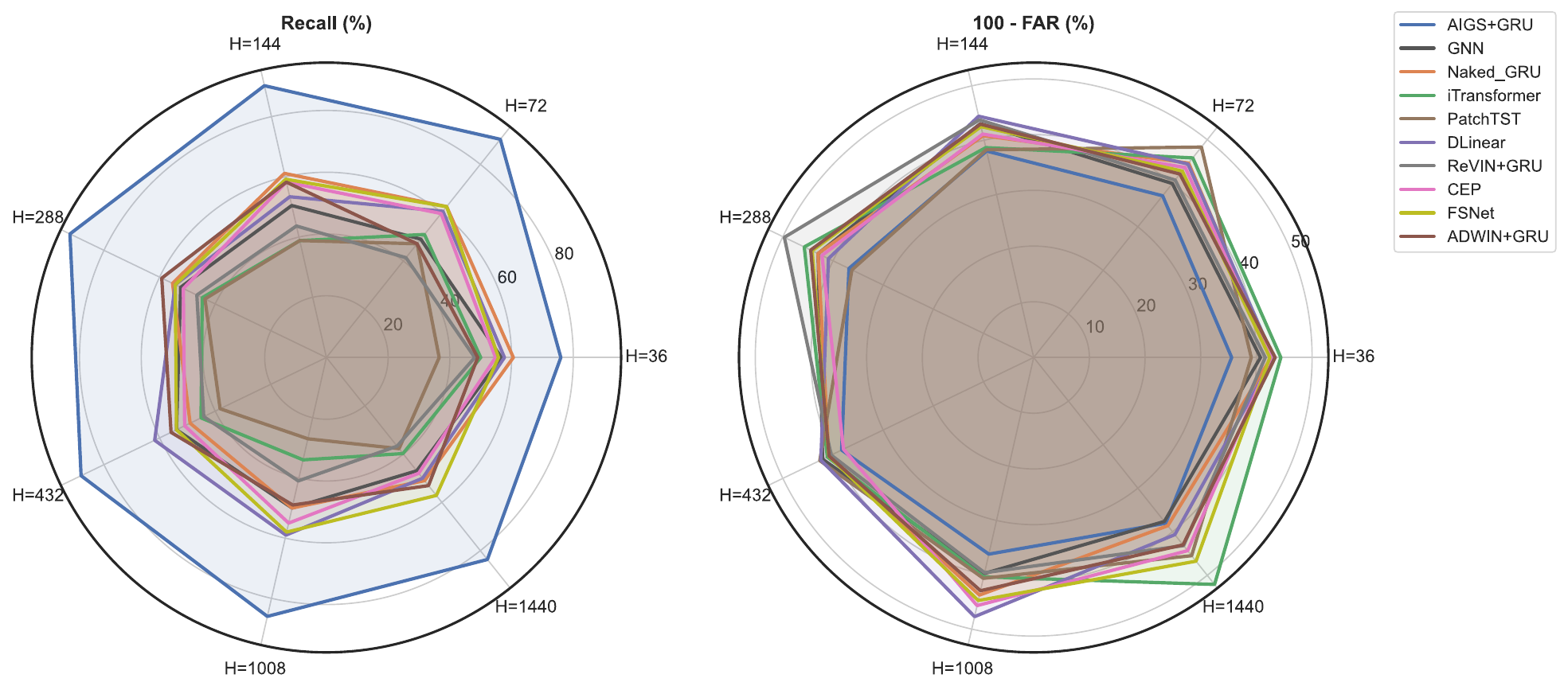}
\caption{Weather Radar Performance for Recall and 100-FAR (\%) across Different Horizons.}
    \Description{Radar chart comparing Weather recall and 100-FAR across horizons for AIGS and baseline models.}
    \label{fig_radar_chart}
\end{figure}

\section{Results and Discussion}
\subsection{Main Forecasting Results and Comparative Analysis}

Tables \ref{tab_pems_results}--\ref{tab_weather_results} report the forecasting results by horizon. The best value is bolded, the second best value is double-underlined, and the next best value is single-underlined. With AIGS placed before the GRU, the system shows a balanced operational profile in streaming environments. AIGS+GRU is not designed to unilaterally dominate heavy deep models on point-forecasting accuracy (RMSE). Instead, the goal is Pareto optimality for edge resource allocation. While massive transformer-based baselines may achieve marginal RMSE advantages in specific short-horizon settings, AIGS+GRU remains highly competitive. Crucially, AIGS demonstrates its true value through domain-dependent state-awareness behavior: providing preventative warning leads on ETT and significantly faster post-shift elastic recovery on PEMS, achieved with only a fraction of the computational footprint.

\subsection{Domain-Specific Adaptation and Proactive Sensing}

Empirical results reveal that AIGS leverages its closed-loop mechanism to exhibit distinct, domain-dependent behavioral advantages. 

ETT is the clearest early-warning case because industrial equipment evolves through gradual, cumulative physical degradation. AIGS+GRU translates minor structural deviations into a preventative warning margin, reaching an $AvgLead$ of 8.31 steps on ETTm1. In contrast, many baseline models (e.g., iTransformer, PatchTST) exhibit negative $AvgLead$ values in Tables 3 and 4. Because these heavy, open-loop models lack immediate feature-space feedback, they blindly extrapolate using historical inertial weights when concept drift occurs, causing their alarms to severely lag behind the actual ground-truth events.

PEMS is dominated by abrupt flow mutations (e.g., sudden traffic congestion) rather than slowly accumulating degradation. In this domain, AIGS fundamentally focuses on post-mutation elastic recovery. AIGS provides the lowest $ART$ across the evaluated horizons. By instantly amplifying plasticity via the Shock Ratio after a sudden mutation, AIGS shrinks the recovery window substantially compared to baseline methods, demonstrating robust elimination of parameter ossification. Figure \ref{fig_efficiency_accuracy} highlights this efficiency-accuracy trade-off, where AIGS+GRU proves optimal for rapid state recovery.

Weather represents a highly noisy environment with low SNR. In such streams, aggressively minimizing point-predictive RMSE often forces heavy models to overfit stochastic random-walk noise. While the RMSE for AIGS is competitive, its true advantage emerges in anomaly recall. The Endogenous Residual Feedback naturally isolates true structural volatility from high-frequency background noise, maintaining robust anti-noise characteristics and capturing extreme climatic events that smoother predictors miss. Figure \ref{fig_radar_chart} shows this recall and false alarm trade-off.

In these experiments, AIGS acts as an internal monitoring signal as well as a feature compressor. Monitoring interpretation is closer to condition-based degradation than to isolated point alarms. In physical systems, structural failures are rarely isolated point impulses; they often appear as drift away from a steady-state manifold before the target variable crosses an event threshold. In that setting, the adaptive gating weight functions as a structural instability indicator and can provide an intervention window. Under abrupt traffic shocks, the same gate is less useful as a long-lead alarm, but it acts as a soft reset-like signal that helps the downstream model reduce obsolete memory after the shift and decreases $ART$.

\begin{table}[tbp]
\centering
\caption{System Resource Utilization and Efficiency Analysis}
\label{tab_memory_report}
\footnotesize
\setlength{\tabcolsep}{2pt} 
\begin{tabular}{lccccc}
\toprule
Model & Time(s) & Peak RAM & Avg RAM & Peak VRAM & Avg VRAM \\
\midrule
Naked\_GRU & 118.30 & 5.1525 & 2.5561 & 160.5005 & 65.7165 \\
Rolling\_PCA+GRU & 133.05 & 4.8944 & 2.2919 & 159.8794 & 65.3334 \\
ReVIN+GRU & 144.61 & 5.2061 & 2.6017 & 161.2690 & 65.3775 \\
AIGS+GRU & 130.94 & 5.7604 & 2.7656 & 160.0381 & 65.8103 \\
\bottomrule
\end{tabular}
\end{table}

\subsection{Resource Efficiency and Edge Viability}
As reported in Table 6, AIGS+GRU takes 130.94 seconds, adding moderate overhead compared with 118.30 seconds for Naked GRU. Average RAM is 2.76 MB and average VRAM is 65.81 MB. Peak RAM is 5.76 MB and peak VRAM is 160.04 MB. These resource profiles explicitly justify the choice of combining AIGS with a lightweight GRU: deploying hundred-million-parameter transformers for real-time stream fine-tuning is engineeringly prohibitive on edge gateways. AIGS proves that through intelligent closed-loop state-aware adaptation, a highly compact $\mathcal{O}\left(k\cdot d\right)$ system can outperform massive models in structural recovery and pre-warning tasks, achieving Pareto optimality in edge deployment.

\begin{table}[tbp]
\centering
\caption{Performance on Different Integration}
\label{tab_linearae_results}
\small
\begin{tabular}{lccc}
\toprule
Model & Recall $\uparrow$ & $AvgLead$ $\uparrow$ & RMSE $\downarrow$ \\
\midrule
AIGS\_CCIPCA+GRU & \textbf{98.8846} & \textbf{11.3024} & \textbf{0.2897} \\
CCIPCA+GRU & 55.2769 & 8.1009 & 0.2926 \\
\midrule
AIGS\_OPAST+GRU & 8\textbf{9.1129} & \textbf{10.6352} & 0.2898 \\
OPAST+GRU & 53.7544 & 7.2879 & \textbf{0.2789} \\
\midrule
AIGS\_LinearAE+GRU & \textbf{93.1284} & \textbf{8.1066} & \textbf{0.2748} \\
LinearAE+GRU & 54.5897 & 7.0291 & 0.2792 \\

\bottomrule
\end{tabular}
\end{table}

\subsection{Generalizability Across Feature Extractors}
The extractor matter is important because AIGS should not depend only on GHA. To test whether the gains come from the gating mechanism itself rather than from isolated synergy with one algorithm, AIGS is evaluated as a model-agnostic framework. GHA is replaced with alternative feature extractors, including CCIPCA, OPAST, and a standard Linear Autoencoder (LinearAE) trained by online gradient descent. Across all feature extractors, integrating AIGS consistently achieves substantial gains in Recall and $AvgLead$ over their standalone counterparts. While predictive accuracy (RMSE) also generally improves, OPAST experiences a slight degradation. This marginal trade-off occurs because AIGS's aggressive plasticity modulation can occasionally destabilize OPAST's recursive internal matrix updates during abrupt structural shifts. In Weather, AIGS-LinearAE reaches 93.13\% recall, compared with 86.68\% for AIGS-GHA and 55.66\% for Naked GRU. It also reaches an $AvgLead$ of 8.11, while AIGS-GHA reaches 7.01 and Naked GRU reaches 7.62. The result should be read as an extractor-agnostic potential in this setting, not a universal independence from the projection module. These consistent improvements across diverse algorithm families demonstrate AIGS's universality and robust generalizability as a plug-and-play architecture.

\begin{table}[tbp]
\centering
\caption{Ablation Study Results}
\label{tab_ablation}
\footnotesize
\setlength{\tabcolsep}{1.5pt} 
\begin{tabular}{lccccccc}
\toprule
Model & ART $\downarrow$ & $AvgLead$ $\uparrow$ & Time $\downarrow$ & MAE $\downarrow$ & $R^2$ $\uparrow$ & RMSE $\downarrow$ & Recall $\uparrow$ \\
\midrule
AIGS(Inc.)+GRU & 9.4662 & \textbf{8.3106} & 16.1222 & 1.0704 & 0.2528 & 1.5068 & \textbf{96.8254} \\
AIGS+GRU (Ours) & \textbf{9.4115} & \textbf{8.3106} & \textbf{15.3366} & \textbf{1.0540} & \textbf{0.2707} & \textbf{1.4886} & \textbf{96.8254} \\
GHA+GRU & 10.6264 & -2.8391 & 12.7210 & 1.1355 & 0.1623 & 1.6015 & 64.7481 \\
Naked\_GRU & 9.7900 & -0.9356 & 16.6441 & 1.1034 & 0.2252 & 1.5359 & 75.9019 \\
Rolling\_PCA+GRU & 10.1623 & -2.5119 & 18.5587 & 1.1078 & 0.2067 & 1.5545 & 72.7273 \\
\bottomrule
\end{tabular}
\end{table}

\subsection{Ablation Study}
An ablation analysis on representative ETTm1 horizons separates the controller components. The "Incomplete AIGS" variant extracts the adaptive weight but lacks downstream error-feedback coupling, while Rolling\_PCA+GRU uses fixed rolling windows. AIGS+GRU reports the lowest ART of 9.41, the lowest time of 15.34 seconds, and the lowest RMSE of 1.49 among the ablation variants while preserving the positive lead time of 8.31. Naked GRU reaches a negative lead time of -0.93, and Rolling\_PCA+GRU degrades further to -2.51. GHA+GRU exhibits the worst lead time and lowest prediction accuracy. The performance gap demonstrates that simple window-based updates are insufficient for capturing the underlying manifold shifts in highly volatile streams. The incomplete AIGS reveals the need for closed-loop learning. While it can be used to trigger warnings with the same lead time, failing to transmit this state to the downstream GRU weakens system-level adaptivity, as indicated by the higher ART value of 9.46.

\begin{figure}[tbp]
    \centering
    \includegraphics[trim=0cm 0cm 0cm 2cm, clip,width=\linewidth]{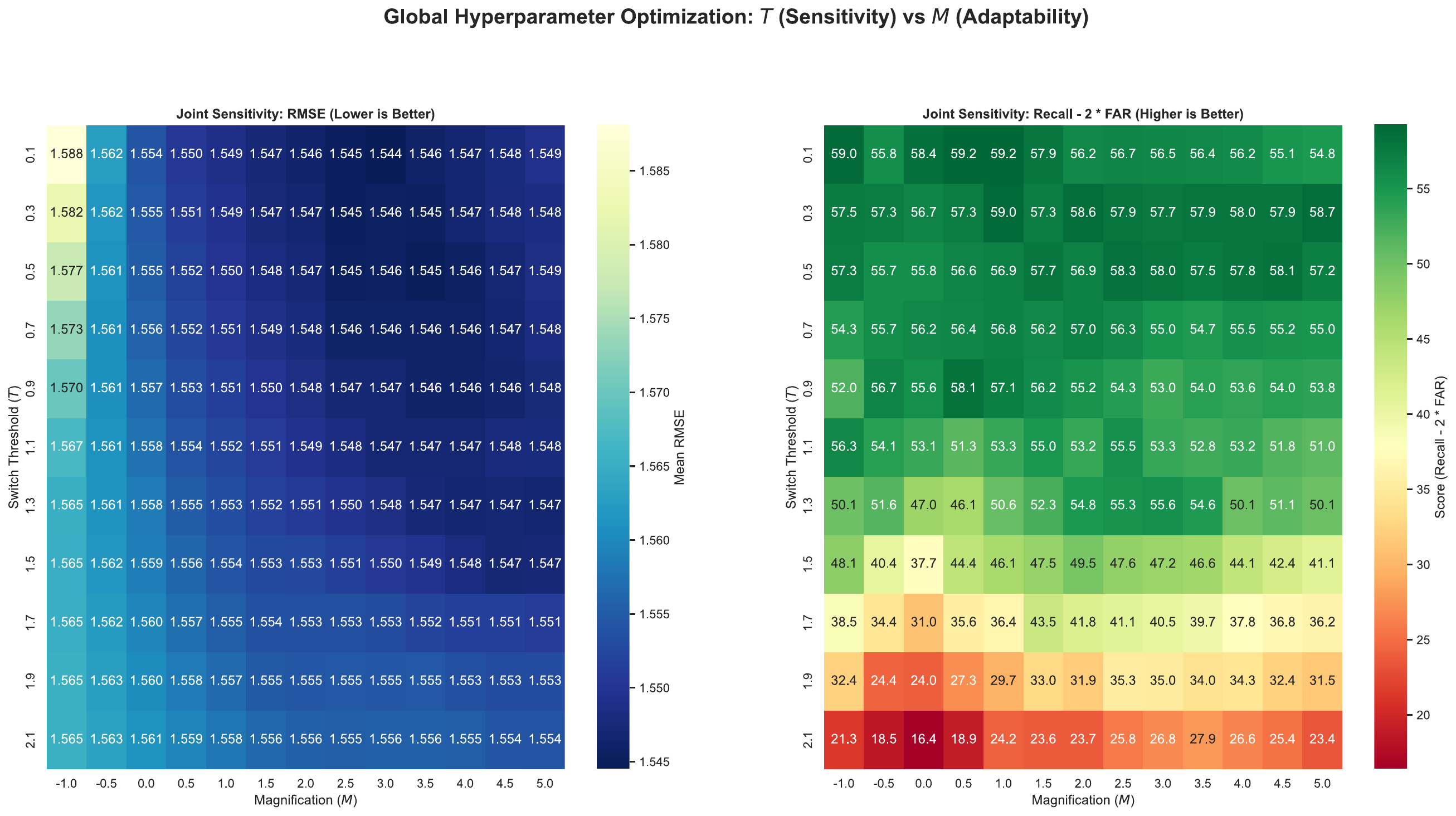}
    \caption{Hyperparameter Optimization for Sensitivity of RMSE and Switching Threshold ($T$) vs Magnification ($M$).}
    \Description{Bivariate surface showing RMSE variation across switching threshold and magnification settings.}
    \label{fig_sensitivity_heatmap}
\end{figure}

\begin{figure}[tbp]
    \centering
    \includegraphics[trim=0cm 0cm 0cm 2cm, clip,width=\linewidth]{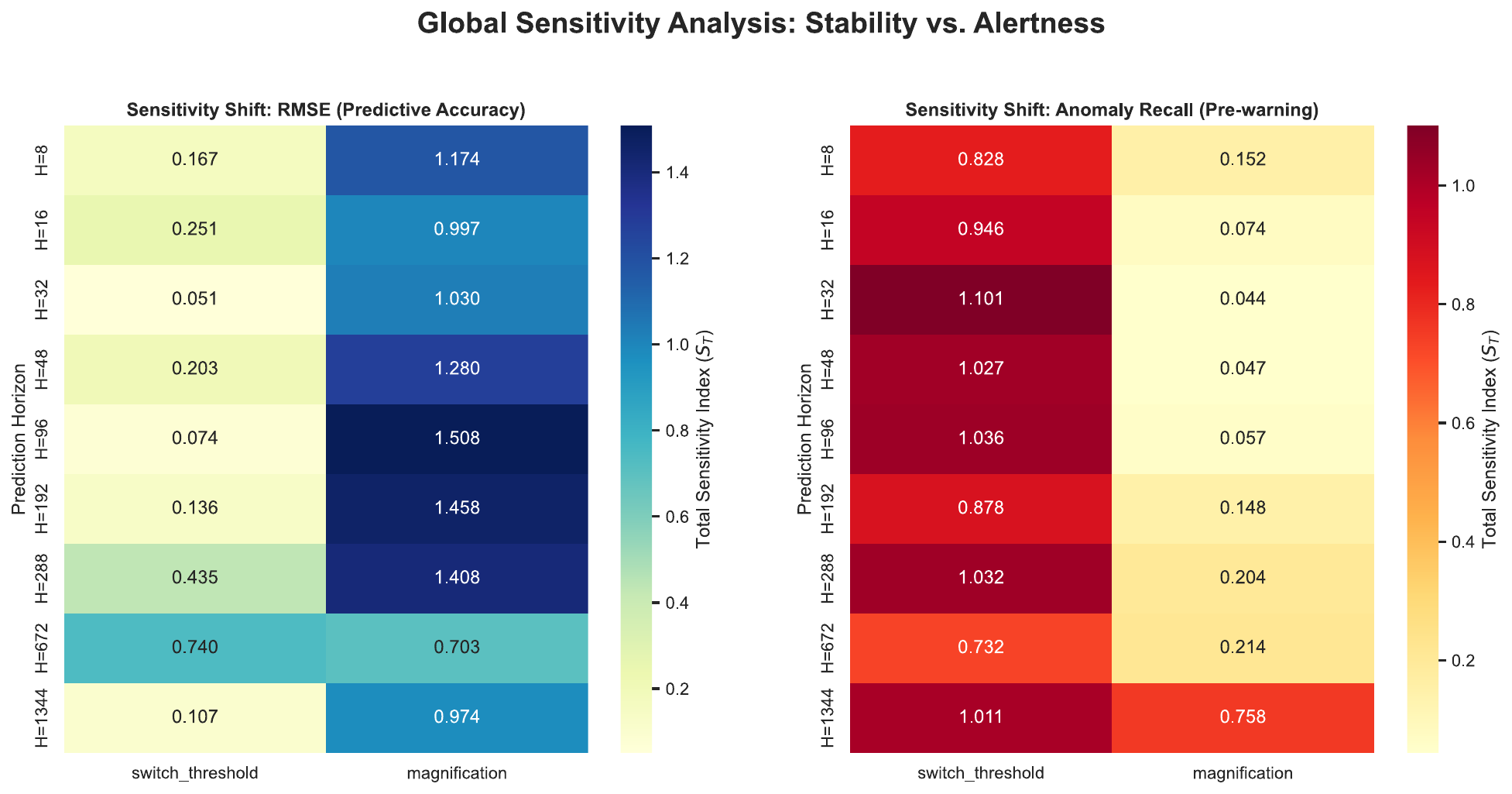}
    \caption{Global Sensitivity Analysis of Stability vs. Alertness across Prediction Horizons ($H$=8 to $H$=1344).}
    \Description{Heatmap showing Sobol sensitivity scores for stability and alertness metrics across prediction horizons.}
    \label{fig_sobol_heatmap}
\end{figure}

\subsection{Hyperparameter Sensitivity Analysis}
To investigate the impact of core hyperparameters on model performance, sensitivity analysis was conducted on the activation threshold $\tau$ and the adaptation magnification $M$. Figure~\ref{fig_sensitivity_heatmap} shows the error surface and the Sobol sensitivity scores.

As observed in the bivariate optimization surfaces, setting $M$ to a relative small value ($M$ $\ge$ 0.0) leave the model with limited plasticity during abrupt shifts. Increasing $M$ beyond the optimal threshold (e.g., $M$ $\le$ 4.5) can push the update toward random fluctuations and degrade RMSE. Except for the highly-noisy Weather, the evaluated datasets, the optimal trade-off is consistently achieved within the ranges of $\tau\in\left[0.5,0.9\right]$ and $M\in\left[1.5,3.5\right]$. Within this optimal region and adjacent suboptimal zones, the baseline predictive accuracy exhibits marginal fluctuations, demonstrating strong architectural robustness against parameter perturbations.

Analysis of the global hyperparameter space demonstrates a
distinct functional decoupling between these two parameters. The Sobol heatmap gives a more direct reading of the parameter space. pre-warning metrics such as $AvgLead$ and Recall are mainly governed by $\tau$. The magnification parameter $M$ has more influence on RMSE, with less influence on the warning trigger itself. The separation supports the practical tuning split between operational stability and state alertness \cite{saltelliVarianceBasedSensitivity2010}.

\section{Conclusion}
In this work, we propose the Adaptive Incremental Gating System (AIGS), a novel closed-loop state-aware adaptation framework for streaming representation learning. To bridge the reconstruction-decision gap in non-stationary streams, AIGS converts underlying reconstruction residuals into Endogenous Residual Feedback. This drives a Continuous Plasticity Controller that smoothly interpolates between learning plasticity and memory retention, ensuring the feature extractor dynamically adapts without catastrophic forgetting. Across WoT and smart city streams, AIGS demonstrates domain-dependent advantages while strictly maintaining $\mathcal{O}\left(k\cdot d\right)$ linear per-step complexity. Empirical evidence highlights its capability to provide proactive early warnings during progressive physical degradation (ETT) and remarkably fast elastic recovery following abrupt flow mutations (PEMS). In highly noisy environments (Weather), AIGS elevates anomaly recall by prioritizing structural volatility over stochastic fluctuations.

While AIGS excels as an adaptive plug-in, its first-order EMA filter inherently requires a brief buffer, prioritizing reliable post-shift recovery over instantaneous spike detection. To overcome current single-channel constraints in multidimensional physical modeling, future work will integrate Graph Neural Networks (GNNs) for spatiotemporal tracking and incorporate covariance alignment to capture higher-order rotational shifts. In conclusion, AIGS provides a robust, lightweight foundation for manifold-aware state monitoring in latency-sensitive edge environments.

\section*{Ethical Considerations}
While AIGS enhances anomaly pre-warning in smart city streams, its deployment requires careful consideration. The empirical trade-off between high anomaly recall and increased false alarm rates (FAR) risks causing operator alarm fatigue in critical infrastructure. Additionally, the model's exponential moving average mechanism introduces a sensing lag, which could delay interventions during abrupt structural failures. Therefore, AIGS should be deployed as an assistive monitoring tool rather than an autonomous decision-maker, enforcing human-in-the-loop verification to balance proactive warnings with operational stability.

\bibliographystyle{ACM-Reference-Format} 
\bibliography{reference}

\end{document}